\documentclass{article}
\usepackage{graphicx}
\usepackage{amsmath} 
\usepackage[preprint]{neurips_2026}

\usepackage[utf8]{inputenc} 
\usepackage[T1]{fontenc}    
\usepackage{hyperref}       
\usepackage{url}            
\usepackage{booktabs}       
\usepackage{amsfonts}       
\usepackage{nicefrac}       
\usepackage{microtype}      
\usepackage{xcolor}         

\title{Zero-Shot Imagined Speech Decoding via Imagined-to-Listened MEG Mapping}

\author{%
  Maryam Maghsoudi\\
  University of Maryland\\
  College Park, MD 20740 \\
  \texttt{maryam00@umd.edu} \\
  \And
  Shihab Shamma \\
  University of Maryland\\
  College Park, MD 20740 \\
  \texttt{sas@umd.edu} \\
}

\begin{document}

\maketitle

\begin{abstract}
  Decoding imagined speech from non-invasive brain recordings remains one of the most challenging problems in brain-computer interface research, as datasets of imagined speech are scarce and difficult to record accurately due to the high variability and uncertainty in the timing and quality of imagined neural activity across subjects and sessions. In this work, we propose a new approach to the decoding of imagined speech that leverages the richer and more reliably labeled recordings during \textit{listening} to speech. We begin by collecting temporally precise Magnetoencephalography (MEG) datasets from a few subjects that paired listened and imagined neural responses to rhythmic melodic, and poem stimuli. All music and speech recordings were made from trained musicians to ensure accurate timing reproductions and alignments (\textit{tempo}) across different subjects and conditions. We then developed a three-stage decoding pipeline inspired by previous findings \cite{marion2021music_imagery} that revealed consistent and meaningful relationships between neural activity evoked by imagining and listening to the same stimuli. In the first stage, we trained a benchmark of six mapping models, employing linear and/or neural network architectures, to map MEG responses during imagining to their listening counterparts. We evaluated these models against a null baseline from unseen subjects to validate that the predicted-listening responses preserve stimulus-specific information. In the second stage, we trained a contrastive word decoder exclusively on the listened MEG responses, and evaluated it using four embedding strategies including semantic, acoustic, and phonetic representations. In the third stage, we process the imagined MEG responses from held-out subjects through the mapping pipeline to compute the corresponding listening responses that are then decoded by the listened decoder. Using rank-based analysis, we show that the imagined words are decodable significantly above chance. We shall report here the results of a proof-of-concept implementation to decode imagined speech, where all evaluations are performed on held-out subjects. We also demonstrate that performance improves with training data size, suggesting that this approach is scalable and can directly be made applicable to realistic brain-computer interface scenarios.
\end{abstract}

\section{Introduction}
Mental imagery is the internal generation of sensory experiences in the absence of external stimuli. It plays an important role in cognitive processes such as planning and memory \cite{kosslyn2001imagery}. Imagined speech refers to internally generated speech without any overt articulation. Studying imagined speech helps understand the fundamental processes of speech generation and also has an important role in  brain-computer interfaces (BCIs) \cite{anumanchipalli2019speech_synthesis}. The ability to decode imagined speech could benefit patients with motor disabilities  such as amyotrophic lateral sclerosis (ALS) \cite{dash2020meg_sensor_selection} or locked-in syndrome (LIS) \cite{ moses2021neuroprosthesis, willett2023speech_neuroprosthesis} who cannot produce overt speech, thus enabling them with other individuals.

One of the main challenges in studying imagined speech is the difficulty of obtaining reliable datasets for training. Unlike overt speech, imagined signals are not easily measurable as it is hard to ensure that a participant instructed to imagine a specific stimulus is actually doing so. Even when the task is performed correctly, there is significant uncertainty in the timing since  we do not know exactly when a particular word or phoneme is being imagined. Furthermore,  imagined speech is often temporally compressed compared to overt speech, and prior work has attempted to address these timing issues in different ways, e.g., by applying dynamic time warping (DTW) \cite{wang2024iad_dtw, martin2016imagined_speech} to account for temporal misalignments, or using rhythmic stimuli such as music, and recruiting trained musicians to better constrain the timing of imagined sequences \cite{marion2021music_imagery, maghsoudi2025imagined_meg_mapping}. Finally, aside from   these experimental challenges, non-invasive imagined neural recordings are typically very noisy, with low signal-to-noise ratios, with unintended artifacts (e.g., eye-blinks)  and substantial variability across subjects. Consequently, collecting large imagined speech datasets is difficult \cite{lopezbernal2022imagined_speech_review}, limiting our ability to  train robust decoding models. These challenges are already  reflected in the decoding results achieved thus far: existing work is often restricted to classification of a small number of words or categories \cite{milyani2025inner_speech, alharbi2024decoding}, such as command words or phonemes, where decoding imagined speech  achieves far lower accuracy compared to listening-based decoding, especially for non-invasive recordings. In addition, many studies evaluate performance \textit{within} subjects rather than across unseen subjects, raising concerns about the decoding generalizability \cite{csaky2025inner_speech_meg_eeg, carvalho2024dda_imagined_speech}.

By contrast to imagined speech, the decoding perceived (listened) speech has seen remarkable progress in recent years.  Studies using invasive approaches such as electrocorticography (ECoG) \cite{moses2021neuroprosthesis, pasley2012reconstructing_speech, akbari2019towards} or stereotactic EEG (sEEG) \cite{he2025vocalmind} have been able to successfully reconstruct speech spectrograms and even intelligible audio. Furthermore,  non-invasive recordings of brain activity  have been shown to be also decodable using linear \cite{gwilliams2022neural} and deep neural networks \cite{defossez2023decoding_speech_noninvasive, tang2023semantic_decoding, liu2026mindmix}, and that decoding performance improves with  scaling of the databases \cite{ozdogan2025libribrain}.  These results suggest that neural responses to listened stimuli contain rich, reliable, and decodable representations and highlight the need for large-scale datasets to train such models.

\begin{figure}
  \centering
  \includegraphics[width=1\linewidth]{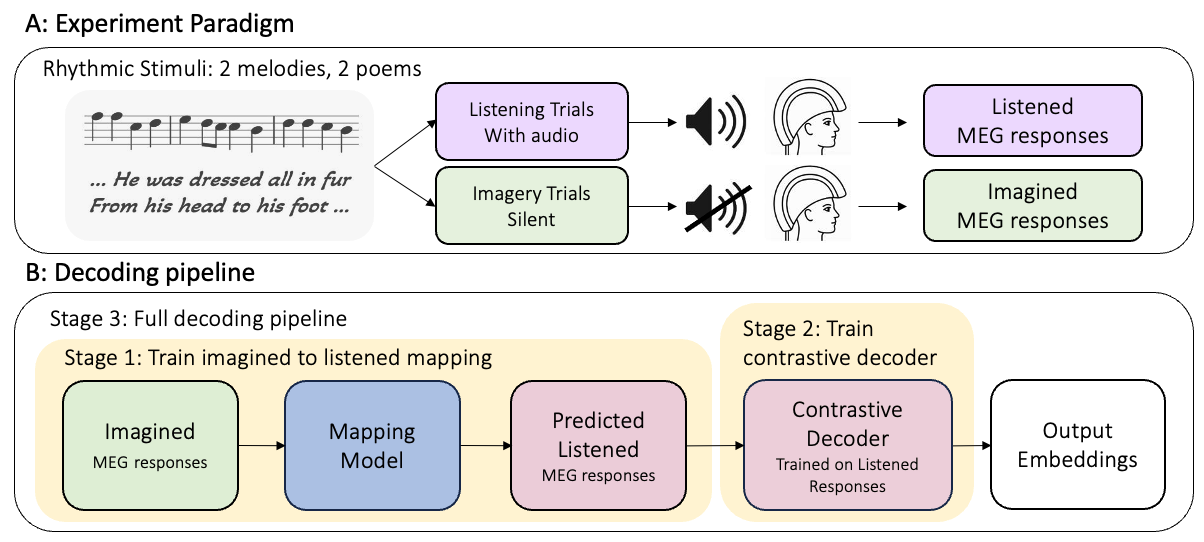}
  \caption{Experiment paradigm and decoding pipeline. \textbf{(A)} Trained musicians were presented with rhythmic stimuli (two melodies and two poems) under two conditions: listening trials, in which audio was played aloud while MEG was recorded, and imagery trials, in which the same stimuli were imagined in silence. Both conditions yield condition-specific MEG responses to the same content. \textbf{(B)} The decoding pipeline proceeds in three stages. Stage 1: an imagined-to-listened mapping model is trained to transform imagined MEG responses into predicted listened MEG responses. Stage 2: a contrastive decoder is trained independently on real listened MEG responses to align neural signals with word embeddings. Stage 3: the full pipeline: imagined MEG from held-out subjects is passed through the frozen mapping model to produce predicted listened responses, which are then fed to the frozen decoder to retrieve output word embeddings. No imagined speech labels are used at any stage of training.}
  \label{fig:story}
\end{figure}

A key motivation for our approach is recent evidence that imagined and listened speech  engage shared neural representations \cite{maghsoudi2025imagined_meg_mapping, alonso2025improving_imagined_speech, kraemer2005musical}. For instance, prior studies have shown that auditory and language-related cortical regions are active during both listening and imagery \cite{zatorre2005musical_imagery, herholz2012musical_imagery}. More recent studies suggest that there is in fact a structured relationship between the two  neural responses \cite{marion2021music_imagery, maghsoudi2026relating_neural_representations}. If this relationship can be learned, a decoder trained exclusively with listened responses could, in principle, be applied to imagined speech without requiring any imagined training labels.

Here we address some of the key challenges of imagined speech decoding at both the data and modeling levels. To mitigate timing variability in imagined datasets, we design an experiment using trained musicians and rhythmic, continuous stimuli in the form of spoken poems, rather than isolated single words. This allows for superior temporal alignments between the imagined and listened conditions. We then investigate whether the relationship between imagined and listened neural responses is learnable using a range of architectures, including both linear and nonlinear models, and evaluate the robustness of these mappings across unseen participants, which is an important requirement for real-world BCI applications. Finally, we train a contrastive decoder exclusively on listened responses and apply it to the predicted-listened activity obtained from the mapping models. In this way, we demonstrate a proof of concept for decoding imagined speech without directly training on imagined data, by first mapping to the listened space and then decoding.

Our main contributions are as follows. First, we introduce a paired imagined-listened MEG dataset collected from trained musicians, using rhythmic, continuous stimuli that enable more precise alignment between conditions. Second, we evaluate the robustness of mapping from imagined to listened responses using six different architectures and show that the predicted listened activity is significantly above a null distribution on unseen subjects, with performance improving as training data increases. Third, we show that these predicted listened responses carry meaningful content through correlation-based classification and show that coarse stimulus identity can be recovered from the predicted responses, while fine-grained segment discrimination remains harder. Finally, we train a contrastive decoder on listened MEG and demonstrate that combining it with the mapping enables above-chance word decoding from imagined MEG on unseen subjects, without directly training on imagined data.

\section{Methods}
\subsection{Data and experiment}
\paragraph{Participants.}A total of 17 participants were recruited in accordance with Institutional Review Board (IRB) approval. All participants provided written informed consent and received monetary compensation. Participants self-reported normal hearing. Given that the timing of imagined auditory experiences is critical in imagery paradigms, we specifically recruited musicians, as they can internalize rhythmic stimuli and maintain temporal accuracy during mental imagery.
\vspace{-5pt}
\paragraph{Stimuli.}The stimulus set consisted of four items: two melodic excerpts from Bach chorales (BWV 263 and BWV 354) and two spoken-word excerpts from the poem \textit{A Visit from St. Nicholas} (1823). All stimuli were loudness-matched to ensure consistency across conditions.
\vspace{-5pt}
\paragraph{Experimental design.}The experiment consisted of 8 conditions: 4 listening and 4 imagery; each corresponding to each of the four stimuli (melody 1 listened, melody 1 imagined, melody 2 listened, melody 2 imagined, and similarly for the two poem snippets). Each condition consisted of 10 trials, each lasting 27 seconds, for a total of 80 trials per session that were presented in a randomized order. 
\vspace{-5pt}
\paragraph{Procedure.}Before the MEG recording session, participants were instructed to memorize all four stimuli. A dedicated pre-session test was conducted to verify that each participant had sufficiently internalized the stimuli and could accurately reproduce them in imagined form. During the MEG recording, a clock-shaped visual metronome was presented to help participants maintain temporal precision during imagery trials.
\vspace{-5pt}
\paragraph{MEG acquisition.}MEG signals were recorded using a whole-head KIT (Kanazawa Institute of Technology) system with 157 axial gradiometers. Data were sampled at 1 kHz, with an online 500 Hz low-pass filter and a 60 Hz notch filter applied during acquisition. All recordings took place inside a magnetically shielded room (VAC). Participants lay in a supine position throughout the session to minimize movement-related artifacts. Auditory stimuli were delivered via ER-3A insert earphones, connected via long silicone tubes to ensure electromagnetic compatibility with the MEG environment.
\vspace{-10pt}
\paragraph{Preprocessing.}
Raw MEG recordings were preprocessed using the MNE-Python toolbox\cite{gramfort2014mne}. Stimulus events were extracted from the trigger channel. To ensure data quality, channels with near-zero variance across time were identified as dead channels and removed. The continuous data were then band-pass filtered between $0.1$ and $40$ Hz to remove slow drifts and high-frequency noise. Artifact removal was performed using Independent Component Analysis (ICA)\cite{hyvarinen2000ica}. The continuous MEG data were segmented into epochs ranging from 0 to 27 seconds relative to stimulus onset.





\subsection{Imagined to listened mapping}
\label{methods_mapping}
The imagined-to-listened mapping is done as a channel-wise sequence-to-sequence regression problem. Given an imagined trial $X \in{R^{C \times T}}$, the model predicts the corresponding listened trial $\hat{Y}\in R^{C \times T}$; where $C$ is the number of MEG channels and $T$ the number of time points at each trial at $100 \text{ Hz}$. Each input trial is z-scored per channel before being fed into the model. To ensure robustness and reproducibility, we implement six different mapping architectures that differ in structure, ranging from a linear system to deep sequence models. These models differ in their temporal inductive bias (none, local, multi-scale, or global) and whether they use past or future context of the trial for predictions.
All architectures are parameterized to approximately the same model size, ensuring that performance differences reflect structural properties rather than parameter count. All the models use the same loss function:
\begin{equation}
    \mathcal{L} = MSE(\hat{Y}, Y) + \lambda (1 - r(\hat{Y}, Y))
\end{equation}
where $r$ is the mean per-channel Pearson correlation between predicted and target responses and $\lambda = 0.5$. The value of $\lambda$ has been chosen empirically. 
\paragraph{Linear lag regression}

We fit a ridge-regression model with lagged features to the neural responses. For each imagined trial, we concatenate shifted copies of that trial from $-100$ ms to $+100$ ms, yielding a feature matrix of dimension $T \times C (2 \delta f_s + 1)$, where $C$ is number of MEG channels, $\delta$ is the lag duration in seconds, and $f_s$ is the downsampled sampling rate. We then solve the regularized ridge regression problem to obtain the weight matrix as:
\begin{equation}
    W = (X_{lag}^T X_{lag} + \alpha I)^{-1} X_{lag}^T Y
\end{equation}
The value of the regularization parameter $\alpha$ is selected by cross-validation.

\paragraph{Neural architectures}
The remaining five architectures are a shallow MLP, a dilated CNN (CNN1D), a UNet with skip connections, a bidirectional GRU (RNN), and a 
temporal convolutional network (TCN). These models capture increasingly complex temporal dependencies, from purely spatial transformations 
(MLP) to multi-scale representations (UNet) and full-sequence context (RNN). Full architectural details are provided in 
Appendix~\ref{app_architectures_details}.

\subsection{Mapping evaluation}

We evaluate mapping quality using two metrics: the mean per-channel 
Pearson correlation between predicted and target listened signals, and 
4-way correlation-based classification performed on predicted listened 
responses. All mapping architectures are trained in a 
leave-one-subject-out (LOSO) manner, where for each held-out subject 
the models are trained on the remaining $N-1$ subjects and tested on 
the held-out subject. To establish chance-level performance, null 
models are trained identically but with shuffled trial labels; each imagined trial was paired with an incorrect listened trial during training. We 
additionally perform a data scaling analysis to measure how mapping 
performance varies with training set size, and a correlation-based 
classification to assess whether predicted responses preserve 
stimulus-specific information. Full details of both analyses are 
provided in Appendix~\ref{appendix_mapping_eval}.

\subsection{Listened decoder}
We trained a contrastive MEG-to-word decoder to identify words from \textit{listened} MEG responses. Word-level onset timestamps were obtained using forced alignment of the poem audio recordings to their ground truth transcripts, using WhisperX \cite{bain2023whisperx}, which combines Whisper-based \cite{radford2023whisper} segment boundary detection with a Wav2Vec2 phoneme alignment model \cite{baevski2020wav2vec}. For each word onset, a 1-second MEG window was extracted spanning 200 ms before to 800 ms after the onset, producing paired samples consisting of a MEG segment and its corresponding word label.

The decoder consisted of two encoders trained jointly: an MEG encoder and a word encoder. The MEG encoder consisted of a spatial 1D convolution across sensors, followed by temporal convolutional blocks with batch normalization, GELU nonlinearities, dropout, and dilated convolutions. The final temporal representation was averaged over time and projected into a 128-dimensional normalized embedding space. In parallel, each word was represented using pretrained language or speech encoders, including BERT (semantic) \cite{devlin2019bert}, Whisper (acoustic) \cite{radford2023whisper}, Wav2Vec2 (phonetic) \cite{baevski2020wav2vec}, and a concatenated $\text{BERT} + \text{Wav2Vec2}$ (combined semantic and phonetic) representation. These pretrained embeddings were passed through a learned projection head to produce normalized 128-dimensional word embeddings (Figure \ref{fig:main_lis_dec}A).






The MEG and word encoders were trained jointly using a symmetric NT-Xent contrastive loss\cite{chen2020simclr}.
\begin{equation}
\mathcal{L} = \tfrac{1}{2} \big[ \mathrm{CE}(\mathbf{S}, \mathbf{I}) + \mathrm{CE}(\mathbf{S}^\top, \mathbf{I}) \big], \quad \mathbf{S}_{ij} = \frac{\mathbf{z}_i^{\mathrm{MEG}} \cdot \mathbf{z}_j^{\mathrm{word}}}{\tau}.
\end{equation}
where $\mathbf{S}_{ij}$ is the scaled dot-product similarity between the $i$-th MEG embedding and the $j$-th word embedding, $\tau = 0.07$ is the temperature, $\mathbf{I}$ is the identity label matrix, and $\mathrm{CE}(\mathbf{S}, \mathbf{I})$ denotes the cross-entropy loss treating each row of $\mathbf{S}$ as logits with the corresponding diagonal entry as the positive class.

Performance was evaluated using rank-based retrieval. For each MEG window, the trained MEG encoder produced an embedding, which was compared against embeddings of all words in the vocabulary. The vocabulary consisted of 76 unique content words drawn from the two poem conditions. The true word rank was computed based on dot-product similarity, where rank 1 indicates that the correct word had the highest similarity. We report rank CDFs compared to chance ($1/\text{number\_of\_words}$). A curve above the chance diagonal indicates that the decoder retrieves the correct word at higher-than-chance rates across all rank thresholds.

\subsection{Full decoding pipeline}
At inference time, imagined MEG segments from a held-out subject are passed through the frozen mapping model, producing predicted listened MEG responses. These are then passed through the frozen contrastive decoder, which compares the resulting embedding against all 76 vocabulary word embeddings and ranks them by dot-product similarity. We evaluate performance using the same rank-based analysis as above. Neither the mapping model nor the decoder has seen any data from the 
held-out subject at any stage of training, making the pipeline zero-shot with respect to imagined speech. To further validate the results, we perform a word consistency analysis measuring whether the 
same words are reliably decoded across subjects and mapping architectures, and whether they align with the words most accessible to the listened decoder. Full details are provided in Appendix~\ref{appendix_consistency}.





\section{Results}
\begin{figure}
  \centering
  \includegraphics[width=\linewidth]{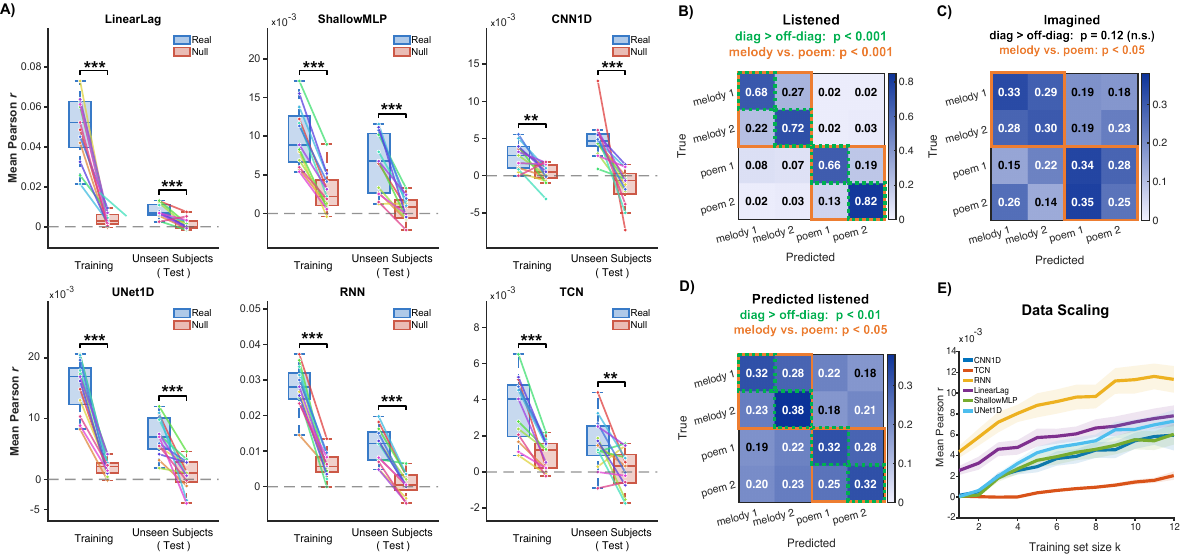}
  \caption{Imagined-to-listened MEG mapping results. 
    \textbf{(A)} Mean Pearson $r$ between predicted and actual listened responses for each mapping architecture, shown under two evaluation conditions: Training and Unseen Subjects (LOSO). Each dot represents one held-out subject, connected lines show paired real-null values. All models significantly outperform their corresponding null baselines (paired $t$-test, $p < 0.001$, $p < 0.01$). Note that y-axis scales differ across models. 
    \textbf{(B)} Confusion matrix for listened MEG responses using correlation-based classification. The diagonal 
    (within-class correlation) is significantly greater than the 
    off-diagonal (between-class correlation) for both the 4-class test 
    ($p < 0.001$) and the coarser melody-versus-poem 2-class test ($p < 0.001$). 
    \textbf{(C)} Confusion matrix for imagined MEG responses. While diagonal entries are not significantly higher than off-diagonal ($p = 0.12$), melody vs. poem classification remains significant ($p < 0.05$).  
    \textbf{(D)} Confusion matrix for predicted listened responses obtained from imagined MEG via the mapping model, using an ensemble voting scheme across architectures, showing recovery of class structure. Both the 4-class ($p < 0.01$) and melody-versus-poem ($p < 0.05$) distinctions are significant
    \textbf{(E)} Data scaling analysis showing performance as a function of training set size $k$ . All models improve with increasing data, indicating that the mapping benefits from additional 
    subject data. Shaded regions indicate 
    standard deviation across held-out subjects.
    }
  \label{fig:main_mapping}
\end{figure}

\subsection{All models learn a significant mapping, with RNN showing 
strongest generalization}
\label{main_mapping_results}
All six mapping architectures achieve prediction correlations significantly 
above null on training data (Figure~\ref{fig:main_mapping}A, $p < 0.01$ for all 
models), confirming that a learnable relationship exists between imagined and 
listened MEG responses. Among the models, LinearLag achieves the highest 
absolute training correlation, yet the RNN shows the largest separation from 
null ($t = 13.02$; Table ~\ref{tab:app_mapping_stats}). This distinction 
suggests that the linear model captures variance that is shared between real 
and shuffled conditions, such as low-frequency drift or session-level 
autocorrelation, whereas the RNN learns features that are more specifically 
tied to stimulus identity.

When generalizing to entirely unseen subjects (LOSO evaluation), all six 
models remain significantly above null (Figure \ref{fig:main_mapping}A), demonstrating that the learned mappings are not subject-specific and transfer to new individuals. The RNN, as shown in Table \ref{tab:app_mapping_stats}, again shows the strongest cross-subject performance ($t = 9.59$; Table ~\ref{tab:app_mapping_stats}), 
followed by ShallowMLP and LinearLag. Notably, the strong performance of LinearLag suggests that a substantial portion of this mapping is linear in nature~\cite{marion2021music_imagery}. 
a simple weighted combination of lagged channel signals is sufficient to 
capture most of the cross-subject transferable structure. The RNN's 
stronger separation from null nonetheless indicates that recurrent 
processing adds something beyond what a linear model can capture, even 
if the difference in absolute correlation is modest. A transformer-based mapping model was also evaluated but did not outperform the null baseline, likely due to the limited dataset size (Appendix~\ref{appendix_transformer}).

\subsection{Mapping predictions preserve stimulus-specific information}
\vspace{-1pt}
To assess whether the predicted listened responses retain stimulus 
identity beyond raw correlation, we applied a correlation-based 4-class 
classification to the predicted signals. When predictions from all models 
are combined via voting, classification accuracy is significantly above 
chance in both the training and LOSO conditions ($p < 0.01$), indicating 
that stimulus-specific structure is preserved in the predicted responses 
even when the model has never observed data from the test subject. At the 
individual model level, four of the six architectures show significant 
classification accuracy (Figure~\ref{fig:app_cm_all}); CNN1D and TCN do 
not reach significance, consistent with their non-significant performance 
on unseen trials of seen subjects (Figure~\ref{fig:app_mapping3}).

To contextualize these results, we compare classification accuracy of the 
predicted listened responses against those of actual listened and imagined 
signals (Figure~\ref{fig:main_mapping}B--D). Actual listened responses 
classify with high accuracy ($72\%$, $p < 0.01$), while actual imagined 
responses classify at only $30\%$ and do not reach significance 
($p = 0.10$), indicating that raw imagined MEG is too noisy to reliably 
discriminate individual stimuli. In a coarser 2-class scheme (melodies 
versus poems), imagined classification is significant ($p < 0.05$), 
revealing that imagined responses carry categorical but not 
stimulus-specific information. By contrast, predicted listened responses 
achieve $34\%$ 4-class accuracy ($p < 0.01$). Although the absolute 
difference from imagined is small, the imagined responses alone are not 
significantly above chance whereas the predicted responses are, indicating 
that the mapping recovers stimulus-discriminative structure that was not 
reliably detectable in the imagined signal.

Additionally, data scaling analysis showed that performance improves monotonically with training set size for all six architectures (Figure~\ref{fig:main_mapping}E), with no model showing 
signs of saturation at the largest training set size tested. This 
suggests that the mapping quality is currently limited by data quantity 
rather than model capacity, and that collecting additional paired imagined 
and listened recordings is a direct path to improving downstream decoding 
performance.

\subsection{Listened contrastive decoder}
\begin{figure}
  \centering
  \includegraphics[width=1\linewidth]{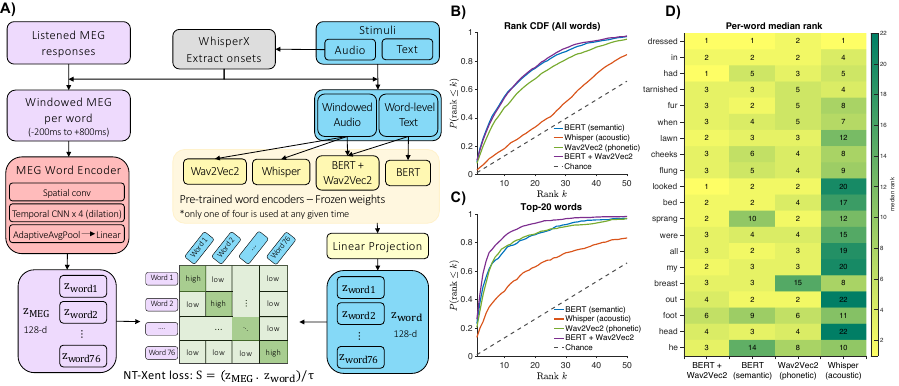}
  \caption{Word-level decoding of listened MEG responses using contrastive 
learning. 
\textbf{(A)} Overview of the decoding pipeline. Continuous listened MEG responses are segmented into word-aligned windows using stimulus onsets extracted with WhisperX. Each MEG segment is encoded into a fixed-dimensional representation using a neural encoder with spatial and temporal convolutional layers. In parallel, corresponding word-level stimuli (audio or text) are processed using pre-trained encoders (Wav2Vec2, Whisper, BERT, or a concatenated $\text{BERT} + \text{Wav2Vec2}$ model) with frozen weights. Both modalities are projected into a shared embedding space and trained using a contrastive loss to align MEG and stimulus representations. \textbf{(B)} Rank-based decoding performance (CDF of $\text{rank} \le k$) across all words, showing the probability that the correct word embedding is 
ranked within the top $k$ candidates. All conditions exceed chance, with BERT and the combined BERT\,+\,Wav2Vec2 encoder performing best. \textbf{(C)} Rank performance restricted to the top-20 best-decoded words (i.e., words with lowest ranks), showing improved performance across all models, with BERT+Wav2Vec2 achieving the strongest results. \textbf{(D)} Per-word median rank for the selected top-20 words across encoder types. Lower values indicate better decoding performance. The multimodal $\text{BERT} + \text{Wav2Vec2}$ representation consistently achieves lower ranks across words.}
  \label{fig:main_lis_dec}
\end{figure}

We trained the listened decoder using a contrastive learning approach 
(Figure~\ref{fig:main_lis_dec}A), bringing the embeddings of correctly 
paired MEG segments and words closer together while separating incorrect 
pairs. We evaluated four word embedding strategies spanning 
semantic (BERT), acoustic (Whisper), phonetic (Wav2Vec2), and combined 
semantic and phonetic ($\text{BERT} + \text{Wav2Vec2}$) representations.

Rank-based evaluation shows above-chance decoding performance for all four encoder models. When evaluated across all words 
(Figure~\ref{fig:main_lis_dec}B), $\text{BERT}$ and 
$\text{BERT} + \text{Wav2Vec2}$ achieve the best performance and perform 
similarly to each other. Wav2Vec2 ranks third, and Whisper performs 
least well, exceeding chance only at larger values of $k$. Nevertheless, 
all four models achieve above-chance decoding across the full vocabulary. 
When restricted to the 20 words the model finds easiest 
(Figure~\ref{fig:main_lis_dec}C), $\text{BERT} + \text{Wav2Vec2}$ 
pulls ahead more clearly, and all curves shift substantially above 
chance. Detailed recall metrics for all encoder models are reported 
in Table~\ref{tab:app_encoder_rank}.

Per-word median ranks across the top-20 words are shown in 
Figure~\ref{fig:main_lis_dec}D. $\text{BERT} + \text{Wav2Vec2}$ performs 
consistently well across most words. Two words are particularly 
interesting. The word \textit{breast} is poorly decoded by Wav2Vec2 
(median rank 15) but well decoded by BERT (median rank 3). This is likely because \textit{breast} has phonetically similar 
neighbors that confuse the phonetic model, whereas its semantic 
distinctiveness in BERT's embedding space makes it easy to 
retrieve. Conversely, the word \textit{he} receives BERT's worst 
ranking (median rank 14), likely because short pronouns occupy 
overlapping regions of BERT's semantic embedding space, making 
MEG-to-BERT alignment difficult for this class of words. Wav2Vec2 
partially recovers performance on \textit{he} (median rank 8), and the 
combined model improves it further (median rank 3), suggesting that 
phonetic information provides a complementary signal for words that are 
semantically underspecified.

\subsection{Imagined speech decoding via the full pipeline}
\begin{figure}
  \centering
  \includegraphics[width=1\linewidth]{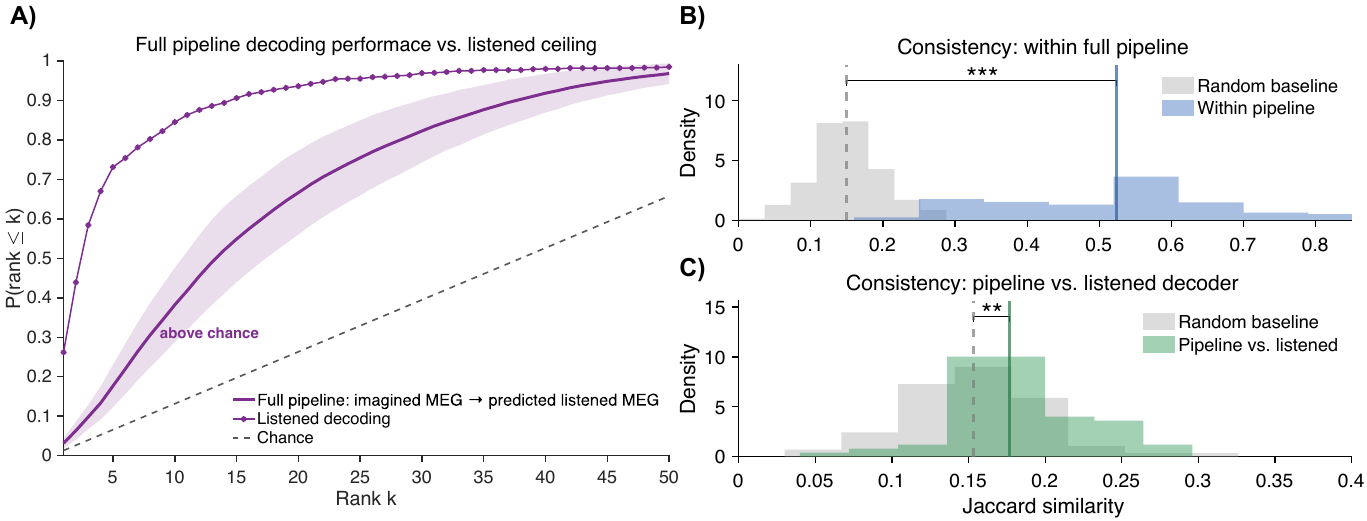}
  \caption{Full pipeline decoding performance and word consistency analysis. \textbf{(A)} Rank CDF for word decoding in our full pipeline (solid line, mean ± standard deviation across held-out subjects and mapping architectures), compared to the listened decoding ceiling (markers) and chance level (dashed), showing significant above chance decoding performance. Results are shown for $\text{BERT} + \text{Wav2Vec2}$ encoder model only. Other encoder models behaved similarly (Appendix \ref{appendix_full_pipeline_results}). \textbf{(B)} Histogram of Jaccard similarity between top-20 best decodable word sets across all subjects and mapping architectures within the full pipeline (blue), compared to a random baseline (gray). The within-pipeline consistency is substantially above chance (Wilcoxon rank-sum test, $p < 0.001$), indicating that the same words are reliably decoded regardless of which subject and mapping architecture is used. \textbf{(C)} Histogram of Jaccard similarity between each subject and mapping architecture's top-20 set and the top-20 words identified by the listened decoder during training (green), compared to the same random baseline. The overlap is significantly above chance ($p < 0.01$), showing that the words the full pipeline decodes well are not arbitrary but align with those the decoder finds best decodable in the listened condition. Vertical lines indicate distribution means. All evaluations are performed on held-out subjects unseen during training.}
  \label{fig:main_full_pipeline}
\end{figure}
We assembled the full decoding pipeline by composing the frozen mapping 
model and frozen contrastive decoder. For each held-out subject, imagined 
MEG was passed through the mapping model to produce predicted listened 
responses, which were then decoded by the listened decoder with no 
imagined data or labels used at any stage of training.

We evaluate decoding performance using rank-based analysis on the output 
word embeddings. Figure~\ref{fig:main_full_pipeline}A shows the rank CDF 
averaged across all held-out subjects and mapping architectures for the 
$\text{BERT} + \text{Wav2Vec2}$ encoder as the chosen best performing encoder, alongside the listened decoding ceiling and chance level. The full pipeline achieves above-chance word 
decoding across all rank thresholds, demonstrating that imagined speech 
can be decoded without any imagined training labels. The gap between the 
pipeline curve and the listened ceiling reflects the compounding noise 
introduced by the mapping step. Rank CDFs broken down by encoder model 
and mapping architecture are provided in 
Appendix~\ref{app_full_pipeline_results}, and show that above-chance 
decoding holds consistently across all encoder and mapping architecture 
combinations.

All subjects individually show above-chance decoding performance (Figure~\ref{fig:app_sbj_variability}), indicating that the pipeline generalizes reliably across individuals and is not driven by a small subset of subjects, which is an important property for real-world BCI applications.

To further validate that the above-chance result reflects meaningful 
decoding rather than arbitrary retrieval, we performed a word 
consistency analysis (Figure~\ref{fig:main_full_pipeline}B--C). The 
distribution of pairwise Jaccard similarity between top-20 decodable 
word sets across held-out subjects and mapping architectures is 
significantly higher than a random baseline ($p < 0.001$). Furthermore, 
the overlap between pipeline top-20 sets and the top-20 words 
identified during listened decoder training is also significantly above 
chance ($p < 0.01$), confirming that the decoded words are consistent across conditions and align with those most accessible to the decoder in the listened condition. Together, these results demonstrate that imagined MEG can be mapped to listened representations, and that the resulting signals retain sufficient stimulus information to enable word-level decoding, providing a proof of concept for zero-shot non-invasive imagined speech BCIs.
\vspace{-2pt}
\section{Discussion}
This work demonstrates that imagined MEG responses can be mapped to 
their listened counterparts, and that the resulting predicted responses 
retain sufficient stimulus information to enable above-chance word-level 
decoding on held-out subjects without any imagined training labels. Notably, all held-out subjects show above-chance performance with consistent sets of decodable words, an important requirement for BCIs that must generalize to new users without per-subject retraining.

Our results show that the listened decoder performs clearly better with 
semantic word encoding models ($\text{BERT}$ and $\text{BERT} + \text{Wav2Vec2}$) compared to acoustic and phonetic models. However, this difference disappears in the full pipeline, where 
all encoder models perform similarly. We also observe a substantial difference  between our full pipeline performance and  the listened 
decoder ceiling, which can be attributed to two sources of loss. First, although the mapping predictions significantly outperform the null distribution and preserve stimulus-specific information, the prediction correlation values are small which introduces noise before the decoder is applied. Second, the decoder is trained on a relatively small 
vocabulary, limiting its generalizability. Our data scaling analysis 
shows that performance increases monotonically with training set size, 
suggesting that both sources of loss may well  be addressed with more data 
rather than necessarily through fundamental architectural changes. 

A notable finding of this study is that the linear mapping model 
(LinearLag) performs competitively with far more complex neural 
architectures, suggesting that the relationship between imagined and 
listened neural responses has a linear structure, consistent with prior 
findings~\cite{marion2021music_imagery}. From a practical standpoint, 
linear models are computationally more manageable, more interpretable, 
and easier to deploy in real-world BCI applications. Additionally, the 
small performance differences across all six mapping strategies suggest 
that the limiting factor is data quantity rather than model 
expressiveness. More powerful architectures such as transformers typically require far more training data to outperform simpler baselines, and our scaling analysis suggests that all models continue to benefit from additional training data. A preliminary evaluation of a transformer-based mapping (Appendix~\ref{appendix_transformer}) confirms this. As larger paired datasets become 
available, transformer-based mapping models and LLM-based decoders are 
a natural next step, and the pipeline introduced here is designed to 
support such extensions.

To reiterate, future work could benefit from scaling the training dataset and extending 
the decoder vocabulary. Additionally, better decoding performance may well result from replacing  the contrastive decoder  with a model trained on larger listened MEG 
datasets which are relatively easy to assemble ~\cite{defossez2023decoding_speech_noninvasive, 
ozdogan2025libribrain}.
The central insight of this work is that the listened speech space, which benefits from large datasets, reliable stimulus alignment, and strong pretrained models can be leveraged to enable imagined speech decoding without requiring imagined labels. As listened MEG datasets 
continue to grow and decoding models improve, the approach introduced 
here provides a scalable and label-efficient path toward practical imagined speech BCIs.

\section{Conclusion}

We present a new approach to imagined speech decoding that requires no imagined training labels. By learning a mapping from imagined to listened MEG responses and composing it with a contrastive decoder trained exclusively on listened data, we achieve above-chance word-level decoding from imagined MEG on held-out subjects. Results are consistent across subjects, mapping architectures, and word encoder strategies and improve with increasing training data, suggesting a clear path toward scaling the approach.


\newpage

{\small
\bibliographystyle{unsrt}
\bibliography{references}
}

\newpage
\appendix
\renewcommand{\thefigure}{A\arabic{figure}}
\setcounter{figure}{0}
\section{Imagery to listening mapping details}

\subsection{Mapping architecture details}
\label{app_architectures_details}
We describe the six mapping architectures evaluated in this work. All 
models take an imagined MEG trial $X \in \mathbb{R}^{C \times T}$ as input and produce a predicted listened MEG trial $\hat{Y} \in \mathbb{R}^{C \times T}$ of the same shape. All architectures are trained with the same loss function and hyperparameter settings described in Section~\ref{methods_mapping}, and are parameterized to approximately the 
same number of trainable parameters to ensure fair comparison.
\paragraph{Shallow MLP}
The input was passed through two fully-connected layers with batch normalisation, GELU activation, and dropout. The input (B,C,T) is reshaped to (BT,C) so that each timestep is processed independently, then reshaped back. This model has no temporal receptive field and it treats each time point of MEG as an independent observation. This model tests whether a purely spatial transformation of the channel space is sufficient for the mapping, without any temporal context.

\paragraph{Dilated CNN (CNN1D)}
We use a convolutional neural network to capture local time dependencies by stacking four depthwise-separable dilated convolutional layers with exponentially increasing dilation factors $d \in \{1, 2, 4, 8\}$. Each convolutional block applies a depthwise convolution that has one filter per input channel, capturing temporal patterns independently per channel, and is followed by a pointwise $1 \times 1$ convolution that mixes information across channels. 

\paragraph{UNet}
The UNet architecture adds an encoder–decoder structure with skip connections to the convolutional backbone. The encoder downsamples the temporal dimension twice using stride-2 convolutions while increasing channel depth. A bottleneck stage processes the most compressed representation with a dilated convolution. The decoder then upsamples back to the original resolution using transposed convolutions, but at each scale it concatenates the corresponding encoder output with the upsampled feature map before applying a further convolution. This concatenation, rather than element-wise addition as in residual networks, allows the decoder to learn separate linear combinations of fine-grained local features (from the encoder) and coarser contextual features (from the bottleneck path).
\paragraph{Bidirectional GRU (RNN)}
The recurrent model processes the MEG sequence with a two-layer bidirectional GRU. A forward GRU to create a hidden state for the past context at each timestep and a backward GRU for future context. The hidden states from both directions are concatenated at each timestep. To keep the total model size comparable to the other architectures, each directional GRU uses a hidden dimension of $h/2$, so the concatenated output has dimension $h$. Dropout is applied between the two GRU layers. Unlike the convolutional models, the GRU maintains a hidden state across the entire sequence, allowing each output timestep to be conditioned on arbitrarily distant past and future context.

\paragraph{Temporal Convolutional Network (TCN)}
The TCN consists of five stacked residual blocks, with dilation doubling across blocks, $d \in \{1, 2, 4, 8, 16\}$. Within each block, two causal dilated convolutions are applied, each followed by batch normalisation, GELU activation, and dropout; the block input is then added element-wise to the block output via a residual connection. The causal structure is enforced by left-padding each convolution by $d (k - 1)$ zeros (where $k$ is the kernel size) and trimming the right end of the output. The TCN uses causal convolutions that would permit real-time deployment in the future.

\paragraph{Parameter counts}
Table~\ref{tab:param_counts} reports the number of trainable parameters 
for each mapping architecture. All neural architectures are within the 
same order of magnitude, ensuring that performance differences reflect 
structural properties rather than parameter count. The LinearLag model 
is not a neural network and its parameter count reflects the number of 
effective regression coefficients ($C \times (2\delta f_s + 1) \times C$).

\begin{table}[h]
\centering
\caption{Trainable parameter counts for each mapping architecture.}
\label{tab:param_counts}
\begin{tabular}{lr}
\toprule
Architecture & Parameters \\
\midrule
LinearLag    & 504,525 (effective) \\
ShallowMLP   & 24,475 \\
CNN1D        & 38,491 \\
UNet1D       & 61,496 \\
RNN          & 57,691 \\
TCN          & 41,787 \\
Transformer  & 120,539 \\
\bottomrule
\end{tabular}
\end{table}

\subsection{Mapping evaluation details}
\label{appendix_mapping_eval}

\paragraph{Correlation-based classification}
To evaluate whether the mapping predictions contain meaningful 
information about stimulus identity, we perform a 4-way 
correlation-based classification on the predicted listened responses. 
Each predicted trial is correlated with the mean listened response for 
each of the four stimulus classes (two melodies and two poems) from the 
held-out subject. The class with the highest mean Pearson correlation 
is assigned as the predicted label. We report the resulting $4 \times 4$ 
confusion matrices and analyze the distribution of diagonal elements, 
comparing them against a null distribution obtained from shuffled 
labels. A significantly higher mean diagonal value indicates that the 
predicted listened responses preserve stimulus-specific information.

\paragraph{Data scaling}
To measure cross-subject generalization as a function of training set 
size, we performed a data scaling analysis. For each held-out subject 
$s$ and training size $k$, we randomly sampled $m = 10$ subsets of $k$ 
subjects from the remaining $N-1$ subjects, trained the mapping models 
on each subset, evaluated on subject $s$, and averaged performance 
across subsets to obtain a stable estimate $r(s, k)$. This procedure 
was repeated for all values of $k$ from 1 to $N-1$, yielding a 
learning curve that shows how mapping performance scales with the 
number of training subjects.

\subsection{Prediction correlation -- Full results}
\begin{figure}[h]
  \centering
  \includegraphics[width=1\linewidth]{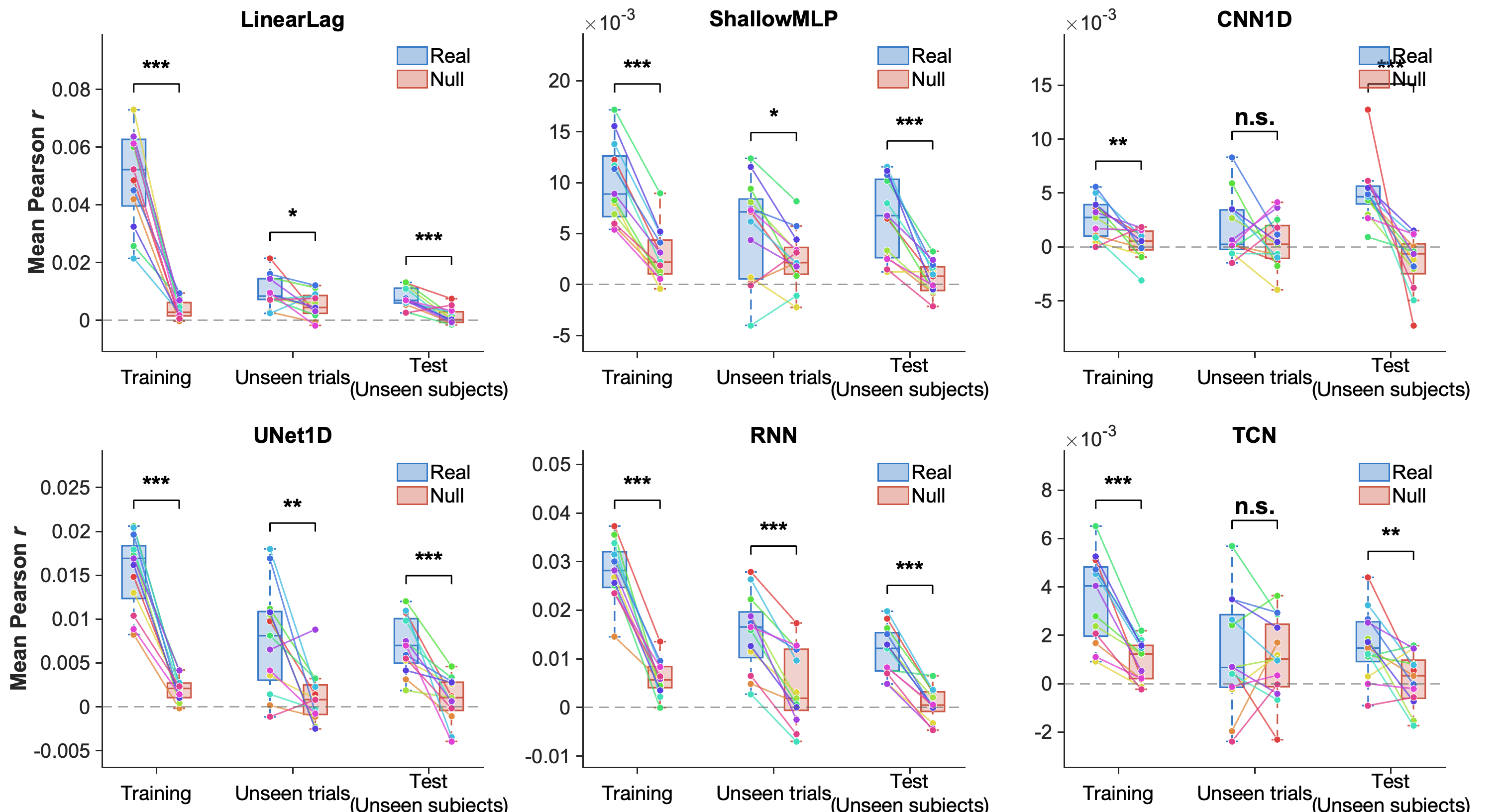}
  \caption{Imagery to listening mapping prediction quality across training, unseen trials, and unseen subjects for all six mapping architectures. Mean Pearson $r$ between predicted and actual listened MEG responses for real mappings (blue) and null distributions (pink), shown across three evaluation conditions: training data, unseen trials from seen subjects, and test data from unseen subjects. Each colored line connects the real and null conditions for a single subject. Significance of the real vs. null difference is indicated by brackets ($*** p < 0.001, ** p < 0.01, * p<0.05$, n.s. not significant). All six architectures show significantly above-null performance during training and test (unseen subjects) conditions. CNN1D and TCN show non-significant differences for unseen trials of seen subjects. The y-axis scales differ across architectures, reflecting differences in the absolute magnitude of predictions.}
  \label{fig:app_mapping3}
\end{figure}

Figure \ref{fig:app_mapping3} reports mean Pearson $r$ between predicted and actual listened MEG responses for all six mapping architectures, evaluated across three conditions: training data, unseen trials from seen subjects, and test data from unseen subjects. Results are shown separately for real mappings and null distributions, with per-subject trajectories connected across conditions. To quantify the statistical significance of these evaluations, we performed a paired t-test on each real vs. null pair (Table \ref{tab:app_mapping_stats})

All architectures maintain significantly above-null predictions even for unseen subjects, supporting the cross-subject generalizability of the learned mappings. RNN shows the greatest significance in the unseen subjects condition (Table \ref{tab:app_mapping_stats}). LinearLag achieves the highest raw $r$ values in training condition.

Two models, CNN1D and TCN, show an anomalous pattern: both are significant in test (unseen subjects) yet fail to reach significance when tested on unseen trials of seen subjects ($p = 0.23$ and $p = 0.80$, respectively). A model 
that generalises to new subjects but not to new trials of the same subject is behaving inconsistently. We interpret this cautiously as evidence that these architectures may be fitting trial-level non-stationarities rather than the true stimulus-response mapping. This pattern warrants attention in future work comparing within- and cross-subject generalization

\begin{table}[h]
  \caption{Paired t-test results comparing real versus null prediction correlations 
for each mapping architecture across three evaluation conditions: Seen (Train): models tested on training sessions of seen subjects, Seen (Test): models 
tested on held-out sessions of seen subjects, and LOSO: models tested on 
entirely unseen subjects. Non-significant results ($p > 0.05$) indicate failure to generalize beyond 
the null model.}
  \label{tab:app_mapping_stats}
  \centering
  \begin{tabular}{lccc}
    \toprule
    Model & Train & Unseen trails of seen subjects & LOSO \\
    \midrule
    LinearLag & $t=10.29,\ p<0.001$ & $t=2.58,\ p=0.024$ & $t=5.63,\ p<0.001$ \\
    ShallowMLP & $t=12.76,\ p<0.001$ & $t=2.83,\ p=0.015$ & $t=6.36,\ p<0.001$ \\
    CNN1D & $t=3.84,\ p=0.002$ & $t=1.28,\ p=0.226$ & $t=4.77,\ p<0.001$ \\
    UNet1D & $t=12.53,\ p<0.001$ & $t=3.67,\ p=0.003$ & $t=5.50,\ p<0.001$ \\
    RNN & $\boldsymbol{t=13.02,\ p<0.001}$ & $\boldsymbol{t=7.68,\ p<0.001}$ & $\boldsymbol{t=9.59,\ p<0.001}$ \\
    TCN & $t=7.18,\ p<0.001$ & $t=0.26,\ p=0.799$ & $t=3.16,\ p=0.008$ \\
    \bottomrule
  \end{tabular}
\end{table}

\subsection{Correlation-based classification details}
\begin{figure}[h]
  \centering
  \includegraphics[width=1\linewidth]{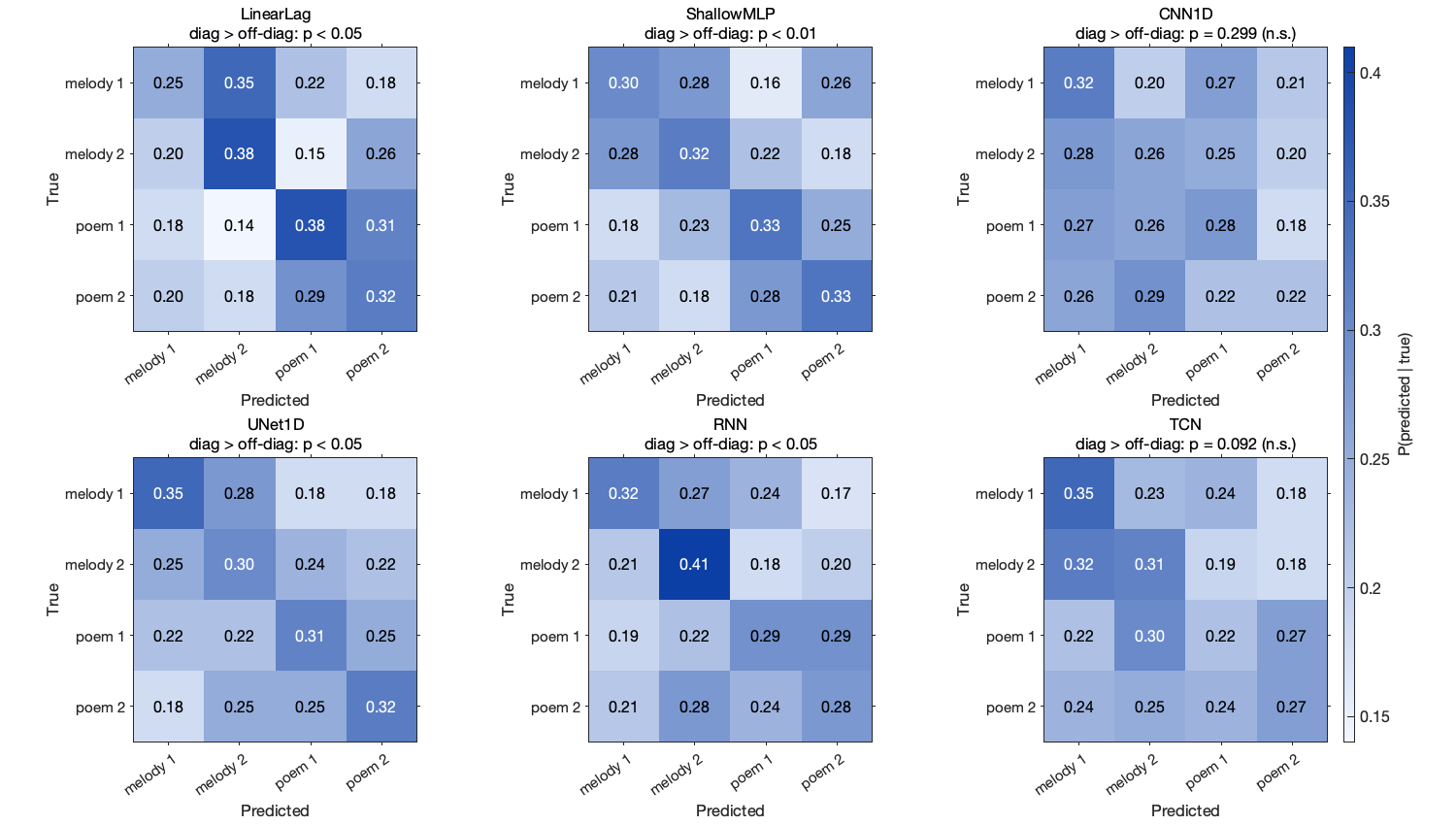}
  \caption{Correlation-based 4-way classification confusion matrices for predicted-listening MEG responses across all six mapping strategies. Each matrix shows the probability of predicting each stimulus category (melody 1, melody 2, poem 1, poem 2) given the true category, computed using correlation-based classification on predicted-listening MEG responses from unseen subjects. Above each matrix, a one-sided Wilcoxon signed-rank test comparing diagonal to off-diagonal values is reported.}
  \label{fig:app_cm_all}
\end{figure}

Figure~\ref{fig:app_cm_all} shows the correlation-based 4-way classification confusion matrices for predicted-listening MEG responses across all six mapping strategies. Classification was performed by correlating each predicted response with the mean listened response for each of the four stimulus categories and selecting the highest. Results are averaged across unseen subjects. Four of six architectures show significantly above-chance diagonal dominance, confirming that the predicted responses preserve coarse stimulus identity. CNN1D and TCN do not reach significance, consistent with their weaker performance in the prediction r-value analysis (Figure~\ref{fig:app_mapping3}).

\section{Transformer-based mapping: motivation and analysis}
\label{appendix_transformer}

A natural question raised by our benchmark of six mapping architectures 
is whether more expressive models, specifically transformer-based architectures, could improve the imagined-to-listened mapping. 
Transformers have demonstrated strong performance across a wide range 
of sequence modeling tasks, and their ability to model long-range 
dependencies via self-attention could in principle be beneficial for 
capturing the full-trial temporal structure of MEG responses. We 
therefore evaluated a temporal transformer as a seventh mapping 
architecture under the same LOSO protocol and training setup as the 
other six models.


\subsection{Results and interpretation}

\begin{figure}[h]
  \centering
  \includegraphics[width=0.35\linewidth]{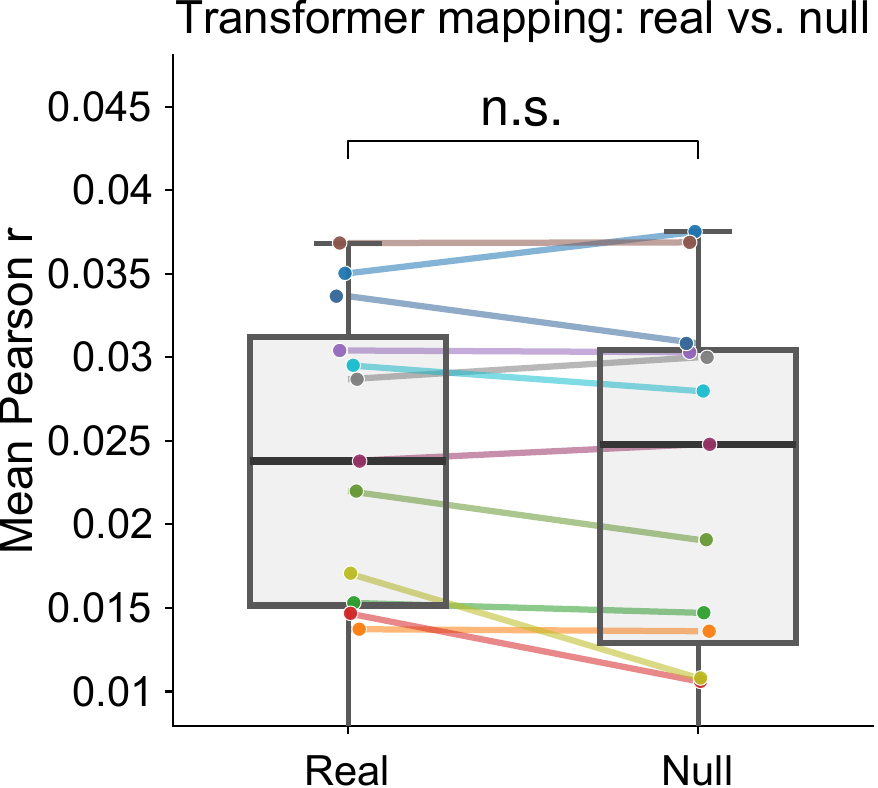}
  \caption{Transformer mapping performance on unseen subjects: real vs. 
null. Mean Pearson $r$ between predicted and actual listened MEG 
responses for the transformer mapping model (real) and the corresponding 
null model trained on shuffled labels (null), shown for each held-out 
subject individually. Colored lines connect real and null values for the 
same subject. The difference is not significant (paired $t$-test: 
$t = 1.71$, $p = 0.114$), indicating that the transformer does not 
learn a mapping that generalizes to unseen subjects under the current 
training setup.}
  \label{fig:app_transformer}
\end{figure}

The transformer did not significantly outperform the null distribution 
on unseen subjects (mean $r = 0.024$ vs. null $r = 0.023$, paired 
$t$-test: $t = 1.71$, $p = 0.114$; Figure~\ref{fig:app_transformer}), 
and 4-class classification accuracy remained at chance level ($25.4\%$ 
real vs. $26.0\%$ null). 

This result suggests that the current dataset size and signal-to-noise ratio are insufficient for effectively training attention-based architectures. Transformers generally require substantially larger datasets than convolutional or linear models because they contain fewer built-in inductive biases regarding locality and temporal structure. In contrast, CNN- and lag-based architectures are naturally well-suited for neural time-series data, where local temporal dependencies dominate. Also, the imagined-to-listened mapping task may itself be largely linear, consistent with the strong performance of the LinearLag model reported in the main paper.

Importantly, this negative result does not necessarily imply that transformer architectures are unsuitable for imagined speech decoding. Rather, it suggests that the current regime is primarily data-limited. Our scaling analysis in the main paper shows that performance consistently improves with additional training data across all architectures. As larger paired imagined-listened datasets become available, transformer-based mapping models may become increasingly advantageous.

\subsection{Architecture}

The temporal transformer takes an imagined MEG trial 
$X \in \mathbb{R}^{C \times T}$ as input and produces a predicted 
listened trial $\hat{Y} \in \mathbb{R}^{C \times T}$ of the same shape. 
It consists of four components. First, a pointwise spatial projection 
($1 \times 1$ convolution) maps the $C = 155$ sensor channels to a 
$d_{\text{model}} = 64$ dimensional model space. Second, sinusoidal 
positional encoding is applied along the time axis to provide temporal 
ordering information to the attention layers. Third, $N = 3$ transformer 
encoder layers with pre-LayerNorm, 4-head self-attention, and a 
two-layer feedforward block ($d_{\text{ffn}} = 128$, GELU activation) 
process the sequence. Full (non-causal) self-attention is used, 
consistent with the offline evaluation setting. Fourth, a linear output 
projection maps from $d_{\text{model}}$ back to $C$ sensor channels. 
The model has 120,539 trainable parameters, comparable to the other 
neural architectures in our benchmark (Table~\ref{tab:param_counts}).

\section{Contrastive decoder training details}

The decoder models were trained with AdamW \cite{loshchilov2019decoupled}, a learning rate of $3 \times 10^{-4}$, weight decay of $10^{-4}$, batch size 64, and early stopping based on validation loss. Gaussian noise was added to MEG windows during training as data augmentation.

\begin{figure}[h]
  \centering
  \includegraphics[width=0.8\linewidth]{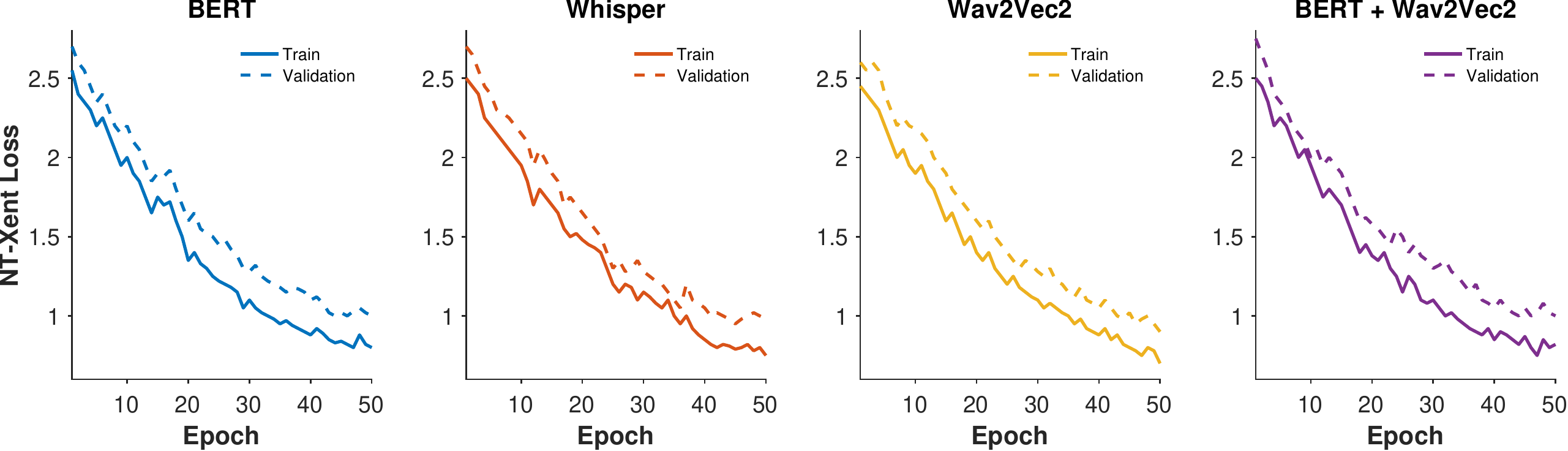}
  \caption{Training curves for the contrastive listened decoder across four word encoder conditions. NT-Xent loss on the training set (solid) and validation set (dashed) over epochs for each of the four word embedding variants: BERT (semantic), Whisper (acoustic), Wav2Vec2 (phonetic), and BERT+Wav2Vec2 (combined). All four conditions converge smoothly without signs of instability. The close tracking between training and validation loss across conditions indicates that the decoder generalizes well within the listened domain and that early stopping was not triggered prematurely.}
  \label{fig:loss_decoder}
\end{figure}

We report training and validation loss curves for each of the four word encoder variants used in the contrastive listened decoder (Figure \ref{fig:loss_decoder}). In all conditions, the MEG encoder and projection head were trained jointly for up to 100 epochs using AdamW with cosine annealing and early stopping based on validation loss. The curves confirm stable optimization across all four embedding strategies, with no evidence of overfitting or divergence. 

We compare word decoding performance across different encoder models on listened MEG (Table~\ref{tab:app_encoder_rank}). Semantic representations (BERT) outperform acoustic ones (Whisper), while phonetic features (Wav2Vec2) provide intermediate performance. Combining semantic and phonetic representations yields the best results across all metrics, indicating that these features capture complementary information for decoding.

\begin{table}[h]
\caption{Word decoding performance on listened MEG across different encoder models. We report Recall@1, Recall@5, and Recall@10. The combined BERT + Wav2Vec2 model achieves the best overall performance.}
\label{tab:app_encoder_rank}
\centering
\begin{tabular}{lccc}
\toprule
\textbf{Encoder} & \textbf{R@1} & \textbf{R@5} & \textbf{R@10} \\
\midrule
BERT (semantic)         & 0.086 & 0.365 & 0.550 \\
Whisper (acoustic)     & 0.030 & 0.113 & 0.183 \\
Wav2Vec2 (phonetic)    & 0.071 & 0.300 & 0.472 \\
BERT + Wav2Vec2        & \textbf{0.091} & \textbf{0.351} & \textbf{0.541} \\
\bottomrule
\end{tabular}
\end{table}

\section{Full pipeline decoding details}
\label{app_full_pipeline_results}
\vspace{-7pt}
\subsection{Decoding rank CDFs details}
\label{appendix_full_pipeline_results}

\begin{figure}[h]
  \centering
  \includegraphics[width=0.65\linewidth]{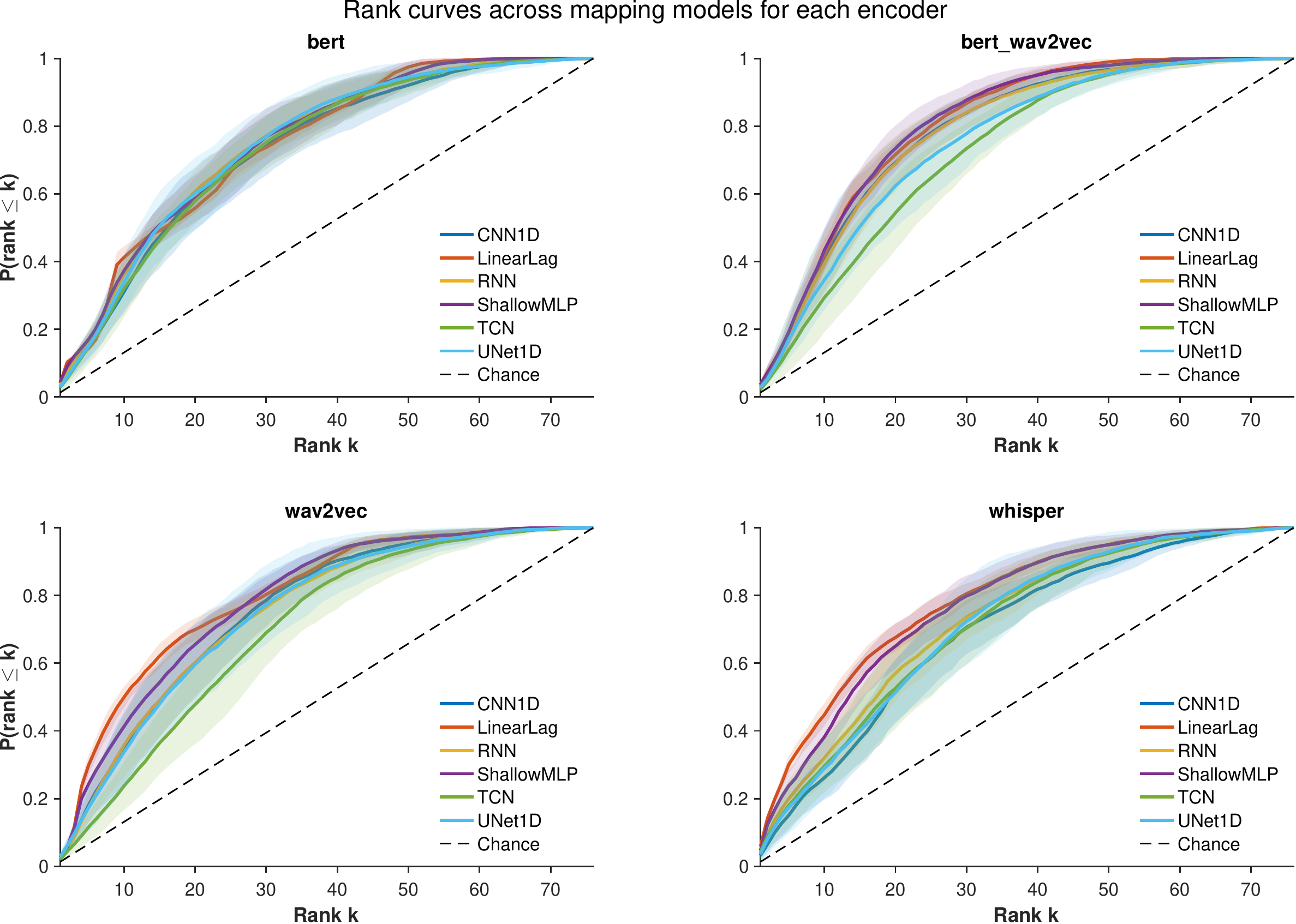}
  \caption{Rank CDFs for the full pipeline decoding across all six mapping 
    architectures, shown separately for each of the four word encoder 
    models (top-20 words, averaged across held-out subjects $\pm$ standard 
    deviation). All mapping architectures achieve above-chance decoding.}
  \label{fig:app_rank_encoders}
\end{figure}

\begin{figure}[h]
  \centering
  \includegraphics[width=0.8\linewidth]{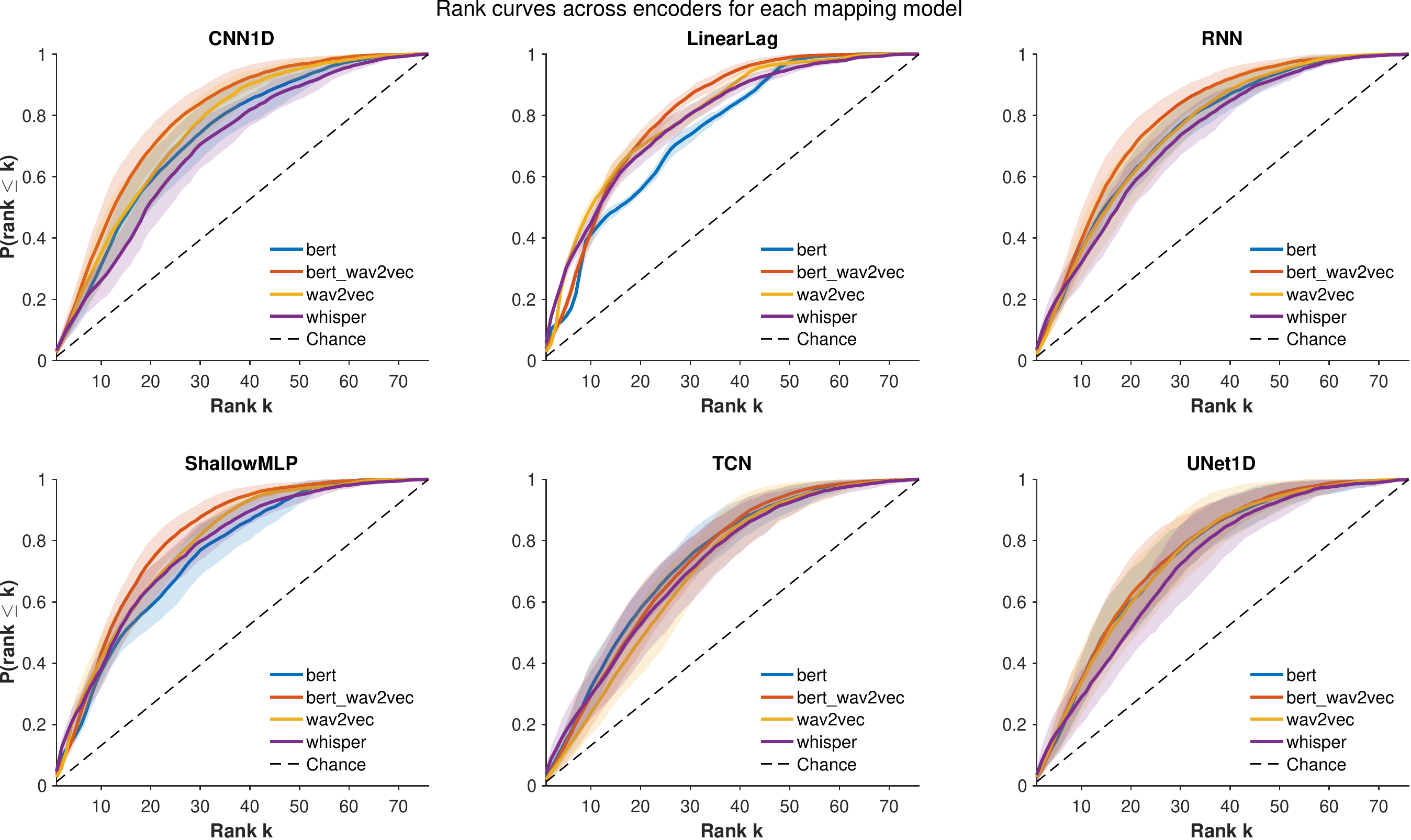}
  \caption{Rank CDFs for the full pipeline decoding across all four word encoder 
models, shown separately for each of the six mapping architectures 
(top-20 words, averaged across held-out subjects $\pm$ standard 
deviation). Above-chance decoding is observed for all mapping 
architectures. The encoder ranking is largely consistent across mapping architectures, suggesting 
that the choice of word representation affects decoding performance 
independently of the mapping model used.}
  \label{fig:app_rank_mapping}
\end{figure}

Figures~\ref{fig:app_rank_encoders} and~\ref{fig:app_rank_mapping} 
show the rank CDFs for the full pipeline broken down by encoder model 
and mapping architecture respectively, evaluated on the top-20 most 
decodable words averaged across held-out subjects. All encoder and 
mapping architecture combinations achieve above-chance decoding 
performance, demonstrating that the result reported in the main text 
is robust and not specific to a particular model choice. $\text{BERT} + \text{Wav2Vec2}$ performs consistently well across all mapping architectures. 

\vspace{5pt}

\subsection{Subject variability}
\label{appendix_subj_variability}

\begin{figure}[h]
  \centering
  \includegraphics[width=0.5\linewidth]{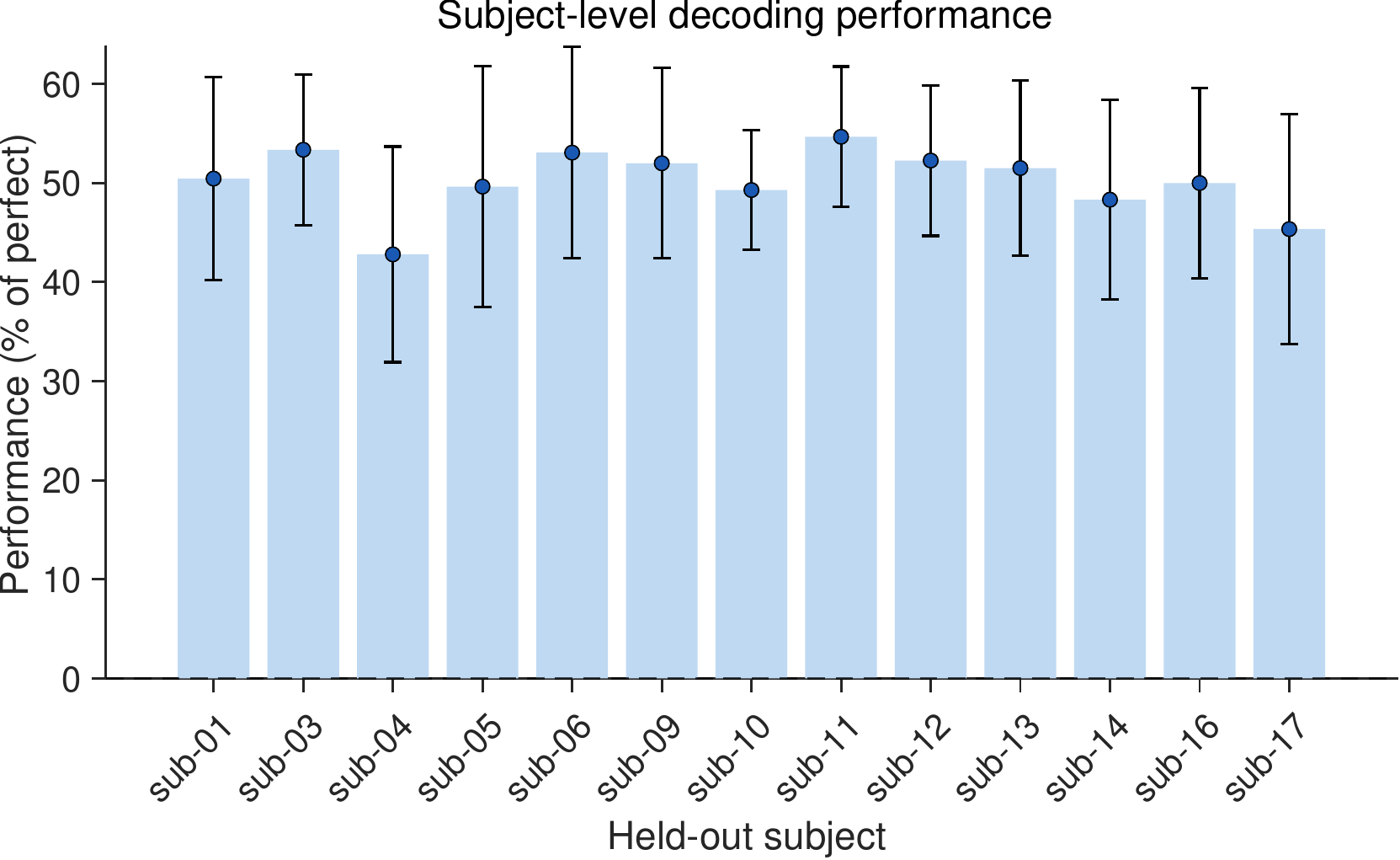}
  \caption{Performance is reported as normalized AUC above chance, expressed as a percentage of perfect decoding. $0\%$ corresponds to chance-level decoding, while $100\%$ corresponds to perfect decoding. Bars denote the mean across all model combinations (6 mapping architectures $\times$ 4 text encoders), with error bars showing standard deviation across combinations. Performance remained consistently above chance across held-out subjects, indicating robust cross-subject generalization.}
  \label{fig:app_sbj_variability}
\end{figure}

Figure~\ref{fig:app_sbj_variability} shows per-subject decoding performance for the complete full-pipeline evaluation. For each held-out subject, imagined MEG was first mapped to predicted listened MEG, passed through the listened-trained decoder, and evaluated using rank-based word decoding. Performance was quantified using the area under the rank curve above chance:
\begin{equation}
\mathrm{AUC}_{\mathrm{above\ chance}}
=
\frac{1}{K}
\sum_{k=1}^{K}
\left[
P(\mathrm{rank} \leq k)
-
\frac{k}{V}
\right],
\end{equation}
where $V$ is the vocabulary size and $K$ is the maximum evaluated rank. The resulting AUC values were normalized by the theoretical maximum possible AUC, yielding performance as a percentage of perfect decoding. Performance remained consistently above chance across all held-out subjects, indicating that the full pipeline generalizes across subjects with relatively limited inter-subject variability.

\subsection{Word consistency analysis}
\label{appendix_consistency}
To assess whether the full pipeline reliably decodes the same words across different subjects and mapping architectures, we identified the top-20 most decodable words for each subject and mapping architecture combination. For each combination, words were ranked by their median rank across all four word encoder models, and the 20 words with the lowest median ranks were selected. We then measured the consistency of these sets of words using Jaccard similarity, comparing all pairwise combinations across subjects and mapping architectures. To contextualize these values, we compared the observed Jaccard similarities against a null distribution obtained by computing Jaccard similarity between randomly sampled sets of 20 words drawn from the full 76-word vocabulary. Additionally, we compared each subject's top-20 set obtained from full pipeline inference against the top-20 words identified by the listened decoder when evaluated on real listened MEG to assess whether the words most accessible to the full pipeline align with those most accessible to the decoder in the listened condition. Statistical significance was assessed using a Wilcoxon rank-sum test.
\newpage

\newpage
\end{document}